# Advancing Melanoma Diagnosis with Self-Supervised Neural Networks: Evaluating the Effectiveness of Different Techniques

Srivishnu Vusirikala
Greenwood High International School
Bengaluru, India
srivishnuvusirikala@gmail.com

Suraj Rajendran B.S.
Weill Cornell Medicine
New York City, The United States of America
sur4002@med.cornell.edu

*Abstract*—We investigate the potential of self-supervision in improving the accuracy of deep learning models trained to classify melanoma patches. Various self-supervision techniques such as rotation prediction, missing patch prediction, and corruption removal were implemented and assessed for their impact on the convolutional neural network's performance. Preliminary results suggest a positive influence of self-supervision methods on the model's accuracy. The study notably demonstrates the efficacy of the corruption removal method in enhancing model performance. Despite observable improvements, we conclude that the self-supervised models have considerable potential for further enhancement, achievable through training over more epochs or expanding the dataset. We suggest exploring other self-supervision methods like Bootstrap Your Own Latent (BYOL) and contrastive learning in future research, emphasizing the cost-benefit trade-off due to their resource-intensive nature. The findings underline the promise of self-supervision in augmenting melanoma detection capabilities of deep learning models.

## I. INTRODUCTION

Melanoma, a prevalent and severe form of skin cancer, poses a significant global health concern, affecting a large number of individuals worldwide. In 2020 alone, approximately 325,000 people were diagnosed with melanoma, with 57,000 fatalities [1]. Given its potential to rapidly progress and become life-threatening, early detection is crucial, as it significantly enhances the effectiveness of melanoma treatment [2]. Over the past decade, remarkable advancements in melanoma treatment have emerged, presenting patients with the opportunity to benefit greatly from these interventions when the disease is identified at an early stage [3].

To optimize the initial weights of neural networks, self-supervision serves as a pre-training technique, surpassing random or non-specialized model weights. Through this process, the input data undergoes modifications, enabling the self-supervised model to learn how to reverse these alterations [4]. Consequently, the model acquires diverse features from the input data, offering a superior starting point for subsequent model training. Various self-supervision techniques are employed today, such as Bootstrap Your Own Latent (BYOL), contrastive learning, rotation prediction, missing patch prediction, and corruption removal, among others [5].

Leveraging machine learning models for early melanoma detection holds immense potential as an affordable and accessible diagnostic tool [6]. However, employing artificial intelligence in the healthcare sector necessitates high levels of accuracy to ensure the avoidance of overlooking potentially hazardous tumors. Therefore, even small enhancements in model accuracy can considerably enhance its utility for the general population. Self-supervision emerges as a highly promising approach for achieving such improvements [7]. Accordingly, this research aims to investigate the impact of self-supervision on the training of a deep learning model for melanoma patch classification.

This study aims to assess the efficacy of self-supervision in enhancing the accuracy of a neural network designed for classifying melanoma patches. Our hypothesis posits that self-supervision will yield accuracy improvements, and the subsequent experiments will either substantiate or refute this proposition.

## II. METHODS

This research focuses on evaluating the effectiveness of various self-supervision algorithms on a convolutional neural network (CNN) trained to detect melanoma patches. Self-supervision, a novel technique for training deep learning models, has gained significant popularity, with models like DINOv2 by Meta gaining attention [8]. Therefore, this study aims to provide evidence regarding the merits of self-supervision amidst the ongoing debate.

### A. Data Availability

A publicly available dataset comprising approximately 10,000 images has been utilized to ensure robust conclusions. The dataset can be accessed and downloaded from: Melanoma Dataset.

### B. Data Generation

The experiment involves three self-supervision methods: rotation prediction, missing patch completion, and corruption removal. Each method employs distinct techniques for generating the input data.

*Rotation prediction data*: Random angles ranging from 0° to 360° are selected, and rotation matrices are generated accordingly. These matrices are then applied to the images to create the input data. Each rotated image is paired with its corresponding angle, forming one training example, as seen in Fig. 1.

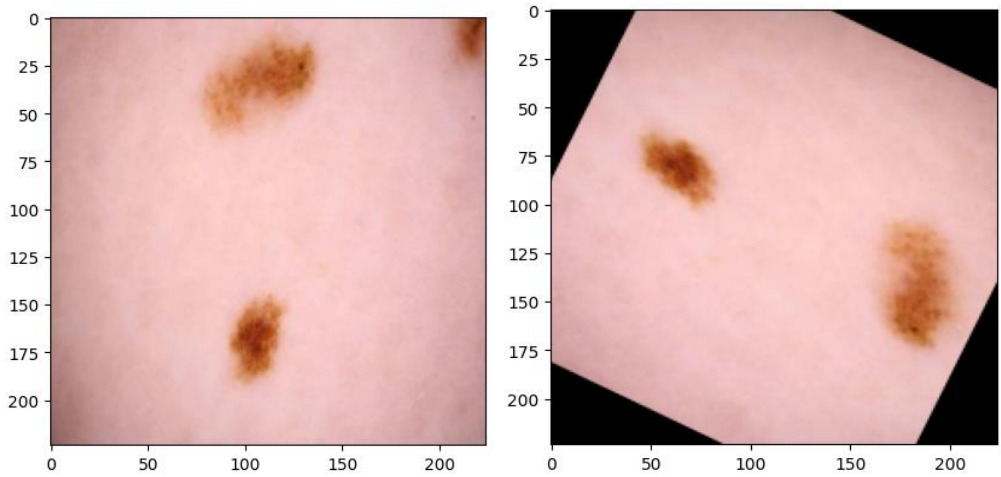
Fig. 1. Original image compared with rotated image

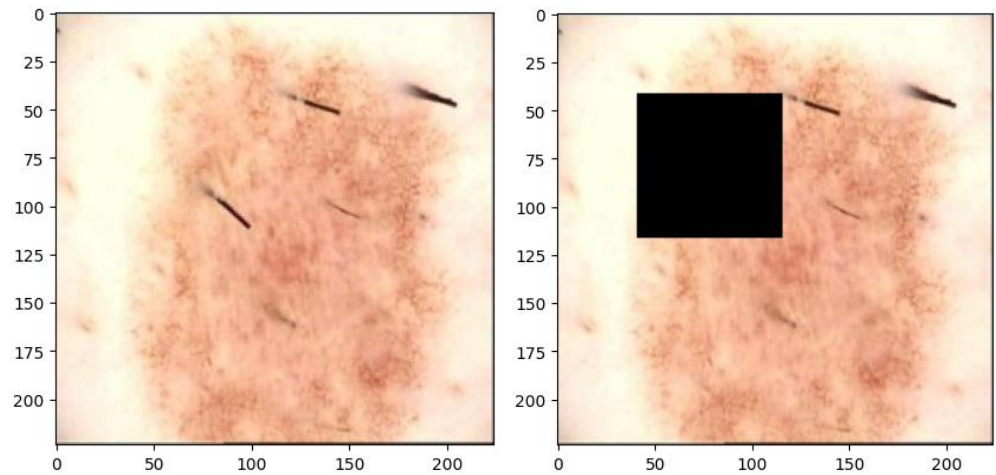
Fig. 2. Original image compared with image with missing patch

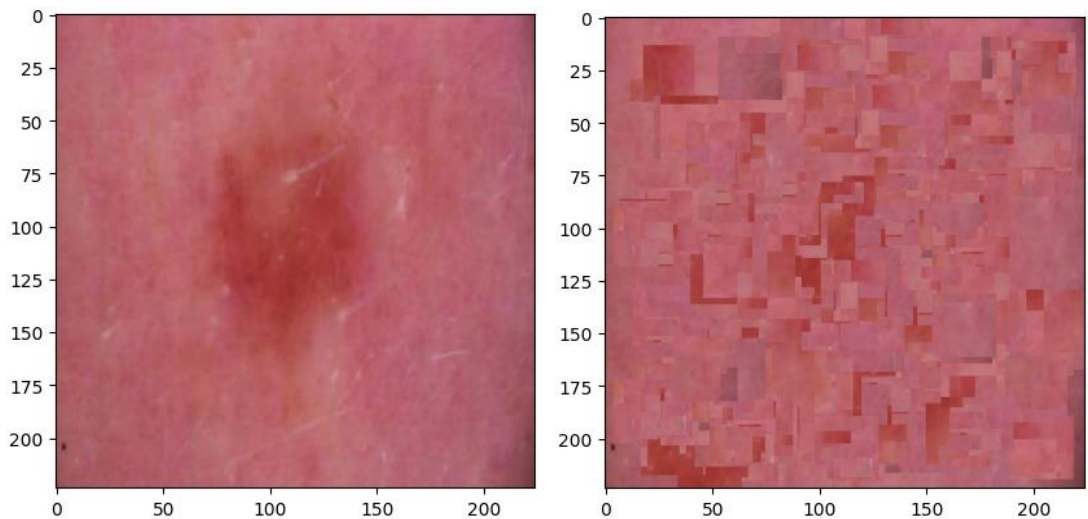
Fig. 3. Original image compared with corrupted image

*Missing patch data*: Square patches with a side length of 75 pixels are randomly removed from the images. A starting coordinate is chosen, and 75 pixels are added to both the x and y coordinates to determine the ending point of the square. The pixels within that region are changed to (0, 0, 0), creating a black patch of 75x75 pixels. The original image, along with the image containing the missing patch, form one training example, as seen in Fig. 2.

*Corruption removal data*: An algorithm is developed to randomly select two patches of 30x30 pixels each and swap their positions. This process is repeated 100 times, resulting in a corrupted image where different patches have been swapped multiple times, making it challenging to discern the original image. This corrupted image is paired with the uncorrupted image, forming one training example, as seen in Fig. 3.

## C. Model Development

The models in this experiment are created using the Keras module of the TensorFlow package [9]. All models share a common base: A Residual Neural Network (ResNet50) utilizing pre-trained ImageNet weights, serving as the encoder. The encoder is then combined with a decoder, with each self-supervised setup having a distinct decoder architecture.

Rotation prediction model: The decoder consists of a Sequential model with four layers: A GlobalAveragePooling2D layer, a Dense layer with 1024 neurons that each utilize a rectified linear unit (ReLU) activation, a Dropout layer with a rate of 0.5, and a Dense layer with a linear activation and a single neuron, serving as the final output layer.

Missing patch model: The decoder employs a Deconvolutional model comprising five sets of Conv2D layers and UpSampling2D layers. The Conv2D layers adopt a (3, 3) filter size and ReLU activation. The number of neurons in each Conv2D layer reduces exponentially from 256 to 16. The output layer is another Conv2D layer with three neurons, a (3, 3) filter size, and linear activation.

Corruption removal model: The decoder architecture is identical to that of the missing patch model, as both models aim to generate an uncorrupted image as the final output.

## D. Model Training

Before training the models, a preprocessing algorithm is applied to the input data, as required for any model utilizing a ResNet architecture. The TensorFlow-provided preprocessing function applies the following changes:

- Conversion of pixel values from RGB to BGR.
- Zero-centering of each color channel.

The models are then trained on the preprocessed images using an Adam optimizer with a learning rate of 0.01, Mean Squared Error loss, and a 0.2 validation split. Training is conducted for 100 epochs.

## E. Model Inference

Following initial training, the models are tested on separate test data to evaluate their performance. Test data specific to this task is generated, and model outputs are compared with the actual values to fine-tune the models for improved accuracy. The missing patch and corruption removal models require an additional step during prediction to obtain a comparable image. A function is implemented to reverse the preprocessing changes, ensuring that unmodified images can be used for comparison.

## F. Model Comparison

After training, the newly trained weights are used to initialize a new ResNet model, which is further trained to predict the presence of melanoma in the input patches. Three such models are trained to evaluate the previously mentioned three self-supervised techniques. Their efficacies are compared against a separate ResNet model solely utilizing ImageNet weights. As all other features remain constant across the models, and as such any disparities in prediction can be attributed to self-supervision. Classification models are evaluated based on their accuracies on the test data, while self-supervision models undergo evaluation using metrics such as Mean Squared Error, Standard Deviation, Average Absolute Difference, and Structural Similarity Index. The collected data is then used to draw conclusions regarding the efficacy of self-supervision.

## III. RESULTS

After conducting the experiments as described in the methods section, data was collected for each of the different models.

For the classification models, accuracy levels were measured on test datasets consisting of 100 images each. The following formula was employed to calculate the accuracy:

$$100 - \frac{\sum_{i=1}^{m}|\hat{y}_i - y_i|}{m} \times 100$$

For the self-supervision models, performance levels were evaluated using mean squared error, standard deviation, structural similarity indexes for the missing patch and corruption removal models, and average absolute difference for the rotation prediction model [10]. The following formulae were utilized for these calculations:

Mean Squared Error:

$$MSE(y, \hat{y}) = \frac{1}{m}\sum_{i=1}^{m}(\hat{y}_i - y)^2$$

Standard Deviation:

$$\sigma = \sqrt{\frac{\sum_{i=1}^{m}\left(|\hat{y}_i - y_i| - \frac{\sum_{k=1}^{m}|\hat{y}_k - y_k|}{m}\right)^2}{m}}$$

Structural Similarity (for missing patch and corruption removal):

$$SSIM(y, \hat{y}) = \frac{(2\mu_y\mu_{\hat{y}} + c_1)(2\sigma_{y\hat{y}} + c_2)}{(\mu_y^2 + \mu_{\hat{y}}^2 + c_1)(\sigma_y^2 + \sigma_{\hat{y}}^2 + c_2)}$$

Average Absolute Difference (for rotation prediction):

$$AAD(y, \hat{y}) = \frac{1}{m}\sum_{i=1}^{m}|\hat{y}_i - y_i|$$

Using these formulae, the results were obtained for all trained models. Firstly, the accuracies of the classification models were recorded. Four models were trained for each run: one without self-supervision, one with rotation self-supervision, one with missing patch self-supervision, and one with corruption self-supervision. Three different training methods were employed for the classification: 100 epochs training, 100 epochs training with an Early Stopping callback on the validation loss with a patience level of 3, and 100 epochs training with an Early Stopping callback on the validation loss with a patience level of 10. The collected results for these models are presented in Table 1.

| Model type | No self-supervision | Rotation prediction | Missing patch | Corruption removal |
|---|---|---|---|---|
| **100 epochs; no Early Stopping** | 88.5% | 88.7% | 88.2% | 89.7% |
| **100 epochs; Early Stopping (patience level: 3)** | 74.1% | 82.9% | 87.6% | 90.2% |
| **100 epochs; Early Stopping (patience level: 10)** | 79.0% | 86.7% | 88.3% | 90.3% |

Table 1. Data collected from the classification task

| Mean Squared Error | Average Absolute Difference (scaled) | Average Absolute Difference (not scaled) | Standard Deviation |
|---|---|---|---|
| 3.83 | 0.062° | 22.3° | 47.1° |

Table 2. Data collected from the rotation prediction self-supervision model

| Self-supervision method | Mean Squared Error (average) | Mean Squared Error (standard deviation) | Structural Similarity Index (average) | Structural Similarity Index (standard deviation) |
|---|---|---|---|---|
| **Missing Patch** | 451.4 | 181.4 | 0.744 | 0.136 |
| **Corruption Removal** | 916.0 | 589.8 | 0.684 | 0.164 |

Table 3. Data collected from the missing patch and corruption removal self-supervision models

Results were also gathered for the self-supervision models themselves. Utilizing the aforementioned formulae, the results were obtained, as outlined in Tables 2 and 3.

## IV. DISCUSSION

The hypothesis of this experiment was that self-supervision would improve the accuracy of neural networks in classifying melanoma patches. The collected data provides strong support for this hypothesis.

While analyzing the data obtained from the self-supervision models, it must be noted that the missing patch and corruption removal methods would have greater potential to enhance the efficacy of the classification model. These techniques allow the models to learn more image features, providing a better foundation for training the classification models on the dataset.

For the rotation prediction self-supervision model, the data in Table 2 reveals a small mean squared error on the predicted angles for the test dataset. This indicates that the model performs well on the test dataset, yielding accurate angle predictions. Additionally, considering the average absolute difference between the predicted angles provides further support to this argument. The model is trained on angles within a range of 0 to 1, enhancing its performance. While the scaled average absolute difference may appear small, the unscaled value offers a more precise depiction of the model's performance. With a value of 22.3°, the model demonstrates a good understanding of the test images, although there is room for improvement. Notably, the dataset exhibits a high standard deviation for the predicted angles, indicating significant variance. Given that the angles were randomly chosen and expected to be similar overall, a large standard deviation is undesirable.

The data from the missing patch model, presented in Table 3, indicates that the model achieves a relatively low average mean squared error when predicting images. Furthermore, the standard deviation of the mean squared error is also observed to be very low, suggesting precise image predictions with minimal deviation from the actual test images. This argument is further supported by the structural similarity index. The average structural similarity index of the predicted images on the test dataset is approximately 0.744, indicating a high level of similarity between the predicted and actual test images. This shows that the model demonstrates a good performance on the test dataset. The low standard deviation of the structural similarity indexes reinforces the notion that the model predicts images with a high level of precision, aligning with the claim supported by the standard deviation of the mean squared errors. Therefore, the data reflects significant learning from the training dataset, while also highlighting the potential for further improvement, as suggested by the average structural similarity index.

The data obtained from the corruption removal model in Table 3 reveals that the model generates predictions with a relatively high mean squared error, particularly when compared to the results of the missing patch model. Additionally, the standard deviation of the mean squared errors is high, indicating greater randomness and lower precision in the predictions, suggesting that the model may not be operating at its optimal capacity. This observation is further supported by the data on the structural similarity indexes. The average structural similarity index of 0.684 is relatively low, indicating a moderate level of similarity between the predicted images and the test images, which falls short of classifying the model as highly effective. However, the low standard deviation of the structural similarity index suggests that the predictions tend to hover around the same average structural similarity index, indicating a degree of precision. Nevertheless, the overall performance of the model highlights significant potential for further improvement, which could lead to better results in the subsequent classification task.

Analyzing the data from the classification task, a superficial observation shows that self-supervision positively impacts the accuracy of the neural network. However, a more detailed analysis reveals additional insights. Training the models without early stopping for 100 epochs shows similar accuracies across all models, with the model without self-supervision even outperforming the missing patch self-supervision model slightly. Overfitting on the training data might have contributed to this outcome. To address this issue, the early stopping callback was employed, aiming to select models with better performance on the test data. With a patience level of 3, the accuracies reduced, largely due to insufficient training time, as the models were trained for only 5 epochs before being stopped by the callback. A patience level of 10 allowed for longer training while still preventing overfitting. This approach resulted in the highest accuracies for the models trained with missing patch and corruption removal self-supervision, surpassing the model without self-supervision. The rotation self-supervision model also exhibited improved performance compared to the model without self-supervision. These results provide further evidence supporting the claim that self-supervision enhances neural network accuracy.

The collected data highlights the potential of self-supervision in improving the accuracy of neural networks classifying melanoma patches. However, there is considerable room for further improvement. Increasing the number of training epochs for the self-supervision models can enhance their accuracy, as evident from the remaining potential for improvement shown in the data. The analysis of the data collected from the classification task indicates that the model utilizing the weights obtained from the corruption removal self-supervision task achieved the highest performance. Moreover, the data collected from the self-supervision model itself demonstrates considerable potential for further enhancement. Therefore, it would be valuable to focus on refining the accuracy of this particular model, as it exhibits the greatest potential for improvement and bears significant influence on the overall classification task. Utilizing a larger training dataset can also be beneficial, allowing the models to gather more comprehensive information and features related to melanoma patches. Moreover, while this experiment focused on three self-supervision algorithms, future research should consider exploring other algorithms like BYOL and contrastive learning, which have the potential for even greater efficiency but require additional time and resources for training[5]. It is important to determine whether the gains in accuracy justify the additional investment. Considering these factors and supported by the data collected in the experiment, self-supervision holds substantial promise for improving the overall effectiveness of neural networks classifying melanoma patches.

## V. ACKNOWLEDGMENT

We would like to acknowledge Lumiere Education for providing the training materials and resources required to conduct parts of this study, as well as providing edits to the manuscript.